\documentclass[final]{ws-rv9x6}
\usepackage[utf8]{inputenc}
\usepackage[square]{ws-rv-van}
\usepackage{xcolor}
\usepackage{graphicx}
\usepackage{hyperref}
\usepackage{amssymb}
\usepackage{mathtools}
\usepackage{ws-rv-thm} 
\usepackage{cleveref}

\newcommand{\CHAPTER}[1]{~\cite{#1}}%% command for arxiv

\begin{document}

\chapter{Distributed Training and Optimization Of Neural Networks}
\label{AIHEP:4.1}

\author{Jean-Roch Vlimant}
\address{California Institute of Technology, Pasadena, CA 91125}
\author[J-R. Vlimant \& J. Yin]{Junqi Yin}
\address{Oak Ridge National Laboratory, Oak Ridge, TN 37831}

\begin{abstract}
Deep learning models are yielding increasingly better performances thanks to multiple factors.
To be successful, model may have large number of parameters or complex architectures and be trained on large dataset.
This leads to large requirements on computing resource and turn around time, even more so when hyper-parameter optimization is done (e.g search over model architectures).
While this is a challenge that goes beyond particle physics, we review the various ways to do the necessary computations in parallel, and put it in the context of high energy physics.
\end{abstract}

\body

\tableofcontents
\section{Introduction}
The main aspects of distributed training have been recently well reviewed in \cite{10.1145/3320060} and we refer to it for a more in depth discussion on technical details.
There exists a rich literature on distributed training of neural networks and notably \cite{10.1145/3320060,10.5555/2999134.2999271,8945109, Kurth2018TensorFlowAS}, recommended as supplementary reading. 

It is commonly agreed that deep learning has shown great success over the last decade, thanks to the creation of large labeled datasets, advancement in model architectures, and increase in computation power --- in part due to general purpose graphical processing units (GPU).
With ever growing complexity of datasets and models, and despite the acceleration provided by GPU, training can still last for days and weeks on single device.
Besides the training of a single model, it is often necessary to perform an optimization over some parameters, that are otherwise not learnable with gradient descent.

With the acceleration of the adoption of deep learning in high energy physics \cite{Radovic:2018dip,Carleo:2019ptp,Bourilkov:2019yoi,livingreview}, it becomes necessary to look at ways to reduce the effective training time.
While most simple neural network models and other classical machine learning methods can be trained in reasonable time, more advanced models like graph neural network \CHAPTER{AIHEP:5.3} and generative adversarial network \CHAPTER{AIHEP:3.1,AIHEP:3.2} can be hard to train~\cite{8638154}.
Complex architectures that exhibit large training time per epoch are often just discarded solely due to the time it would take to bring them to converge --- let alone doing hyper-parameter tuning.
Improvement of the time to solution is required to make the development of such models more amenable.
Distributed training may reduce weeks of training down to days.

It should be noted that the challenge of accelerating the time to convergence is not specific to high energy physics (HEP).
However, the specific computing and software environment of HEP might limit the possibilities otherwise available. 
For example, due to budget constraints, it is not given that GPUs are available for training.
Furthermore, the software options are limited by the requirement of affordable long term support.

\paragraph{}

We provide in this chapter a description of the key aspects of distributed training and optimization as a practical guide to developing large models with large amount of data.
This chapter is organized as follows.
After introducing the formalism of training and optimizing neural network models in section \ref{sec:WF}, highlighting the possible strategies to parallelize computation, we start in section \ref{sec:PS} with the parameter distribution strategy, which was the first to be adopted as a way to speed up the training of models.
We then describe in section \ref{sec:DD} the data distributed strategy that seems to be widely adopted currently, thanks to its ease of use.
We go over model parallelism in section \ref{sec:MP} and recent development in generic deployment of this otherwise complicated method to implement.
In section \ref{sec:HO}, we get into the details of model search and optimization, key to the success of developing high-performance deep learning applications.
We conclude in section \ref{sec:SD} with an overview of advancements of software and outlooks on distributed training. 

\section{Neural Network Optimization Formalism}\label{sec:WF}

Neural network models can generally be represented as a function  $f_{\theta,h}(X)$ predicting some quantity $Y$. $X$ is some input, $\theta$ and $h$ are parameter vectors.
A loss function $\mathcal{L}_{\theta,h} \coloneqq F(f(X))\vert_{\theta,h}$ is defined to characterise the fitness of the model to a specific task --- for example, the binary cross entropy can be used in a binary classification supervised task.
In the general setup, one is searching for the two sets of parameters $\theta^\star$ and $h^\ast$, that provide an optimal value over some data.
We distinguish the parameters $h$ from $\theta$ by the fact that $\mathcal{L}$ is differentiable with respect to $\theta$, but not with respect to $h$ --- $h$ are the so-called \textit{hyper-parameters}.

By definition $\theta^\star$ can be approximated by using the gradient descent method based gradient of the loss function $\mathcal{L}$ with respect to the model parameters $\theta$, noted:
\begin{equation}
    \nabla_\theta{\mathcal{L}_{\theta,h}} = \nabla{F}\vert_{f_{\theta,h}} \bullet \nabla_\theta{f_{\theta,h}}
\end{equation}
starting from an initial set of parameters $\theta_{h,0}$ that may depend on $h$.
We note that some models and loss definitions involve gradient ascent instead of gradient descent and the methods described in this chapter are all applicable in this situation.
In practice, \textit{stochastic gradient descent} (SGD) is found to be more stable than using individual gradients to update the model parameters.
In this algorithm the gradient is averaged over subsets $\{X_{i_b}\}$ indexed by $b$ --- referred to as \textit{batch} therein, but also called \textit{mini-batch} in the literature --- of the whole training dataset.
In such case, the gradient for each individual input can be computed in parallel before calculating the average over a batch as

\begin{equation}\label{eq:data-parallel}
    \nabla_\theta{\mathcal{L}}\vert_b \coloneqq \mathbf{E}_{\{X_{i_b}\}} \left[ \nabla_\theta{\mathcal{L}_{\theta,h}} \right]
\end{equation}
This is the key ingredient that makes the computation of SGD efficient on GPU as it is under single program multiple data (SPMD) programming style.
During the optimization procedure, the parameter $\theta$ is updated with the rule 
\begin{equation}\label{equ:updaterule}
    \theta \leftarrow \theta - \eta \nabla_\theta{\mathcal{L}}\vert_b
\end{equation}
where $\eta$ is either fixed or dynamic and referred to as the \textit{learning rate}.
Equation~\ref{equ:updaterule} can be written more generically
\begin{equation}\label{eq:update}
\theta \leftarrow \theta - \Psi(h,\nabla_\theta{\mathcal{L}}\vert_b)
\end{equation}
where $\Psi(.,x) = \eta x$ in its simplest form of equation~\ref{equ:updaterule}.
An \textit{epoch} is the cycle of the algorithm during which the whole training dataset --- all batches --- is used once to update the model parameters.
It typically requires numerous epochs to train a model properly.
It should be noted that the typical neural network loss function is non-convex and therefore optimization with gradient descent can be subject to being stuck in local minima, preventing the procedure to reach the absolute global minimum --- which might actually be degenerate. 
Obviously, one may compute the effective gradient over a given batch by splitting the computation over multiple parts --- or \textit{shards} --- of a batch.
This is the key aspect of data distribution to which we come back in section \ref{sec:DD}.
Additionally, one may compute the effective gradient of several batches in parallel, with the caveat that applying the update rule (equation~\ref{equ:updaterule}) is not exactly possible (see section~\ref{sec:PS}).% since the gradient computed at a given $\theta$ does not correspond necessarily to the parameter $\theta$ it is applied to.

The case of generative adversarial networks (GAN) brings in some additional complication because of the presence of two models and a training procedure that alternates between updating the parameters of the \textit{discriminator/critic} and the \textit{generator}. 
We come back to these points in section~\ref{sec:PS}.

In the case where the model is composed of multiple layers $L_{l}$ indexed by $l$, i.e. $f_{\theta,h} \coloneqq  L_{N_L}\left( \cdots L_1 \left(  L_0 (X) \right) \cdots \right) \coloneqq \left[ \circ_{l=0}^{N_L} L_{l} \right] (X)$, the derivation chain rule applies and we obtain
\begin{equation}\label{eq:grad}
   \nabla_{\theta_{i}^{m}}{f} = \left\{ \prod_{l={N_L}}^{m+1} {\left. \nabla L_l  \right\vert_{A_{l-1}(X)}} \right\} \left.\nabla_{\theta_{i}^{m}}{L_m}\right\vert_{A_{m-1}(X)}
\end{equation}
where $\theta_i^m$ is a parameter of layer $L_m$. We introduced the notation $A_m(X) = \left[ \circ_{l=0}^{m} L_{l} \right] (X)$ for the \textit{activation} of layer $m$.
The \textit{forward pass} is the computation of the value of successive layers using activation values for preceding layers.
\textit{Back propagation} is the computation of the gradients for layer $L_{m}$ using gradients already computed for any layers $l > m$.

In multiple occasions, the computation required for a single layer $L_l$ and its Jacobian $\nabla_{L_l}$ involve tensor products that allow for some level of parallelism.
Convolutional layers for example can have the computation of each filter on each patch of the input image be processed on separate devices.
We consider this as model parallelism and come back to this point in section~\ref{sec:MP}.

Assuming an arbitrary layer separation $S$ such that $m\leq S<N_L$, the right hand side term of equation~\ref{eq:grad} factorizes as
\begin{equation}\label{eq:model-parallel}
\left[ \left\{ \prod_{l={N_L}}^{S+1} {\left. \nabla L_l  \right\vert_{\left[ \circ_{k=S+1}^{l-1} L_{k} \right] \left( A_S(X) \right)}} \right\} \right] \left[ \left\{ \prod_{l=S}^{m+1} {\left. \nabla L_l  \right\vert_{A_{l-1}(X)}} \right\} \left.\nabla_{\theta_{i}^{m}}{L_m}\right\vert_{A_{m-1}(X)} \right]
\end{equation}
The first bracketed product (referred to as $B_S$) requires only one input involving the layers $m \leq S$ : the term $A_S(X)$. 
Furthermore, $B_S$ appears in the calculation of the gradients for all layers $m\leq S$.
Therefore, the calculations of the gradients for the set of layers $L_{m>S}$ and the set of layers $l_{m\leq S}$ only require knowledge of the two disjoint sets of layers respectively, once $A_S$ (during the forward pass) and $B_S$ (during back propagation) have been computed.
$S$ can hence be chosen to balance memory utilization for example.
This is a key aspect of model parallelism that we expand on in section~\ref{sec:MP}.

The optimal value $h^\ast$ needs to be found with other optimization methods --- notably Bayesian optimization and evolutionary algorithms.
We note that if all values of $h$ are continuous, $h^\ast$ can still be found with gradient descent, using numerical evaluation of the gradient --- with a finite difference method for example.
The discussion on \textit{hyper-parameter optimization} (HPO) in this chapter is unchanged if one decides to use any other non differentiable function as \textit{figure of merit}: $\mathcal{F}_{\theta,h}(\{X_i\})$, instead of using $\mathcal{L}$.
By extension, we include in the hyper-parameter set, all parameters used towards obtaining the optimal set of parameters for the model --- such as learning rate, model initialization parameters, batch size, etc. 
The optimal model parameters obtained at the optimal value of the hyper-parameters $h^{\ast}$ is noted $\theta^{\ast}$.

The full dataset is often divided into three independent subsets: \textit{training}, \textit{validation} and \textit{testing} sets.
The search for $\theta^\star$ (\textit{training} procedure) is done on the training set, the validation set is used to estimate the generalization performance, and the testing set is used solely to report a final performance of the optimal model.
The search for $h^\ast$ (HPO procedure) is done by optimizing $\mathcal{F}_{\theta,h}$ over the validation set, and provides further ways in which the computation may be done in parallel.
The \textit{K-Folding} procedure is recommended when comparing model performance as it provides a better estimation of the mean and variance of the performance.
We come back to the process of hyper-parameter optimization using K-folding in section~\ref{sec:HO}.

\section{Parameter Distribution}\label{sec:PS}

As explained in section~\ref{sec:WF}, the calculation of the gradient averaged over a batch can be done for multiple batches in parallel.
The caveat is that the gradients are calculated from a given set of model parameters and not necessarily applied to update the same model parameters.
To be more precise, given two batches $b_1$ and $b_2$, in the gradient update done sequentially, we have
\begin{equation}
\begin{split}
     \theta_1 \leftarrow \theta_0 - \eta \nabla_\theta{\mathcal{L}}\vert_{b_1,\theta_{0}} \\ \theta_2 \leftarrow \theta_1 - \eta \nabla_\theta{\mathcal{L}}\vert_{b_2,\theta_{1}}
\end{split}
\end{equation}
While we obtain
\begin{equation}\label{eq:stalegrad}
\begin{split}
    \theta_1 \leftarrow \theta_0 - \eta \nabla_\theta{\mathcal{L}}\vert_{b_1,\theta_{0}}  \\  \theta_2 \leftarrow \theta_1 - \eta \nabla_\theta{\mathcal{L}}\vert_{b_2,\theta_{0}}
\end{split}
\end{equation}
in the case of computing the gradients concurrently and applying them sequentially.
On the second update in equation~\ref{eq:stalegrad}, the gradients are computed on $b_2$, from $\theta_0$, but applied on $\theta_1$ --- different than $\theta_0$.
This results in \textit{staleness of gradients} and a slowdown in convergence of the models~\cite{10.5555/2969239.2969316,DBLP:journals/corr/abs-1712-05878}. 
The benefit is that the time to run a full epoch linearly decreases with the number of batches processed in parallel.
The cost is that the decrease of the loss after a fixed number of epochs is not necessarily better when increasing the number of batches processed in parallel.

Multiple proposals have been made to mitigate the effect of outdated gradients, mostly resulting in changing the value of $\eta$.
Having $\eta$ be a matrix in an element-wise multiplication as in the \textit{adagrad} algorithm~\cite{10.5555/1953048.2021068} provides improvements in convergence~\cite{10.5555/2999134.2999271}.
Alternatively, inspired from a physics concept, the \textit{gradient energy matching} algorithm~\cite{DBLP:journals/corr/abs-1805-08469} offers a provable way of stabilizing convergence of the distributed training.

It should be noted that the mode in which the gradients --- that are computed in parallel --- are used to update a the model parameters is subject to contention.
Indeed a bottleneck occurs when the effective time for gradients to be computed is smaller than the update time of the model parameters, if there are many parallel processes for instance.
This underlines the fact that the parameter strategy cannot scale to an infinite number of parallel processes.
One solution to this problem is to create a multi-level~\cite{DBLP:journals/corr/abs-1712-05878} (as opposed to binary level so far) hierarchy of processes that reports updates up one level at a time, effectively reducing the single-updating-process bottleneck, at the cost of increasing the staleness of gradients.

As mentioned in section~\ref{sec:WF} and \CHAPTER{AIHEP:3.1,AIHEP:3.2}, GAN are composed of two models that are trained in an alternate manner, and possibly with different number of parameter updates at each cycle.
The parameters can be synchronized either after both models were trained on a full batch, or after each parameter update, or each model.
The former is the strategy adopted in~\cite{10.1007/978-3-030-02465-9_35,8638154,Vlimant:2019tkb}.
The latter would be impacted even more strongly by the bottleneck created in updating the parameters centrally.

\paragraph{Take Home Message.} 
The method to parallelize training using the parameter distribution strategy is intrinsically limited in scalability due to a combination of the centrality of the parameter update and staleness of the gradients.
Training GAN under such strategy is furthermore complicated by the presence of two models with different training dynamics.
The mild acceleration obtained with parameter distribution can be complemented with other means of parallelization --- as we will see the next sections of this chapter.

\section{Data Distributed Training}\label{sec:DD}

\begin{figure}[h!]
  \centering
  \includegraphics[scale=0.38]{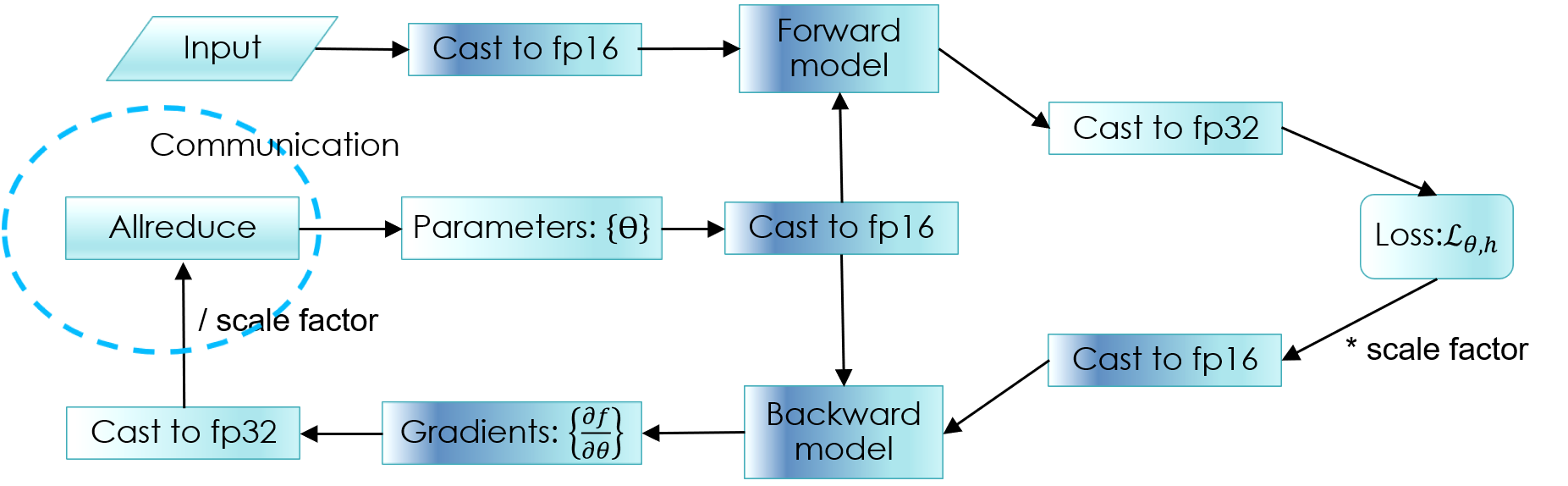}
  \caption{Typical processing flow chart in a data distributed training, where the potential half precision operations (fp16) are marked with mixed colors. The communication is invoked after gradients are estimated from backward model, and the parameters of the model are then updated (Eq.~\ref{eq:update}) with synchronized gradients. The scale factor is needed to avoid the numerical overflow during the $fp16 \leftarrow fp32$ conversion.
} \label{fig:dltrain}
\end{figure}

The data distributed training follows the SPMD paradigm and is the most widely adopted approach to the distributed deep learning due to several advantages that it can offer: straightforward to implement, model agnostic to apply and generally efficient to scale up.
In a typical data parallel paradigm flow chart as shown in Figure ~\ref{fig:dltrain}, a model is replicated on each device and the forward pass on different shards of a batch are calculated independently following Eq.~\ref{eq:data-parallel}.
The gradients of all processes are synchronized after the back propagation, via the \textit{allreduce} collective communication. As the parallelism increases, the effective batch size hence is linearly proportional to the total number of processes (model replicas).
The forward model, backward model, and gradients calculations can often be performed in half precision operations (fp16), and are marked in mixed colors in the flow chart of a typical mixed precision training. The \textit{allreduce} is usually performed in single precision to avoid gradient overflow, but fp16 is also supported by most up-to-date libraries.   
With neural network like ResNet50, where the message size per model replica is about 100 MB, the allreduce communication can soon become a bottleneck when training at scale. 
Based on the novel ring-allreduce algorithm, Horovod~\cite{horovod} is a high performance communication library for data distributed training. 
The algorithm is network bandwidth optimized, and each process sends and receives gradient messages $2(n-1)/n$ times, where $n$ is the number of processes.
It can therefore scale efficiently for large $n$ since the total message size communicated per process becomes a constant when $n \rightarrow \infty$.
Given the success of ever-larger neural network models and data sets, popular deep learning frameworks also provide built-in support for data parallel training. 
The commonly used data-parallel methods in TensorFlow and PyTorch are listed in Table~\ref{tab:dp}.
The advantages of using these methods for distributed training are compatibility (integrated with the framework), user-friendlyness and performance (give the support for high performing communication fabrics).
The usage mode (API) is however not flexible enough to support novel communication patterns and mostly limited to data parallelism only.

\begin{table}[!b]
\centering
\tbl{Framework built-in data parallel support. MirroredStrategy and MultiWorkerMirroredStrategy are equivalent to DataParallel and DistributedDataParallel, respectively.}{
\begin{tabular}{c|c|c}
%\hline
Framework \textbackslash Distribution & Single node      & Multi node                  \\ \hline
TensorFlow                            & MirroredStrategy & MultiWorkerMirroredStrategy \\ \hline
PyTorch                               & DataParallel     & DistributedDataParallel     \\ \hline
\end{tabular} }
\label{tab:dp}
\end{table}

One common pitfall for data parallel training is that model accuracy deteriorates at large batch sizes.
It is argued \cite{large-batch-training} that the estimation of the gradient at each step is more accurate with larger batches and hence the optimization becomes smoother --- i.e. with less stochasticity ---, resulting in a higher probability to be trapped in a local minima.
There are strategies to mediate this effect such as layer-wise adaptive rate scaling (LARS) \cite{lars}, but in general the upper bound for batch size is still limited around $64K$ for most applications based on first-order optimization(e.g. SGD). 
Natural gradient methods can push the boundary of the large-batch training further~\cite{kfac} at the cost of expensive evaluations of second order derivatives.% to better guide the optimization. 

Another pitfall is that batch normalization layers~\cite{batch-norm} are not effective when the shard size is too small, since there are not enough samples to obtain sufficiently accurate mean and variance of the layer activation values.
The synchronization of such layer is not commonly implemented in data parallel libraries (e.g. Horovod, PyTorch DataParallel, etc).
The models with batch normalization layer (widely used after a convolution layer to improve regularization and accuracy), trained with small shard size --- i.e. many processes --- perform significantly worse in both convergence and accuracy, than that of a model trained without data parallelism.
Therefore, the optimal batch for data-parallel training is not simply the optimal batch size for single replica divided by total number of processes ; it depends on both the data and the model.

\begin{figure}[h!]
  \centering
  \includegraphics[scale=0.6]{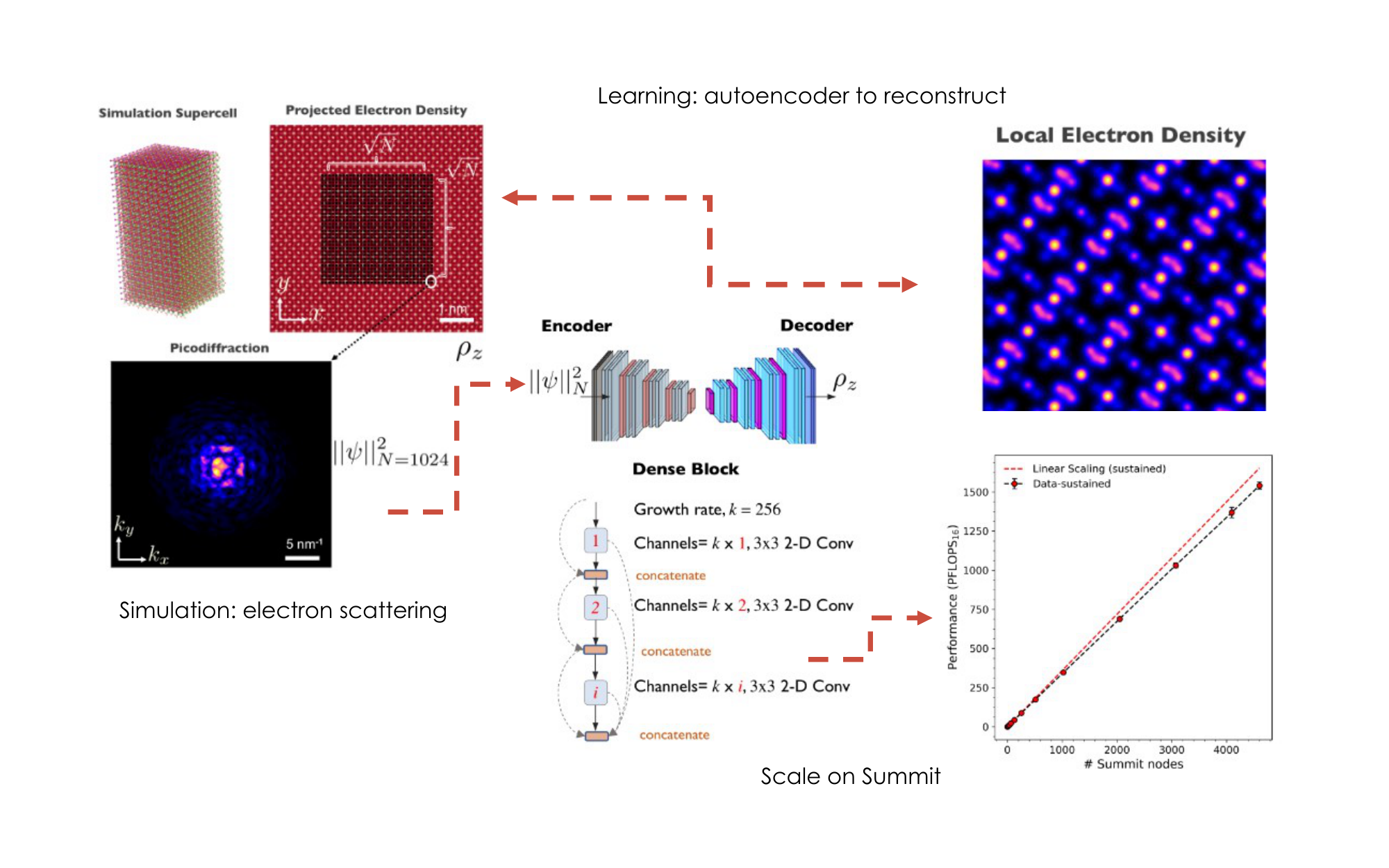}
  \caption{An example \cite{laanait2019exascale} of Exascale data-parallel deep learning on the Summit supercomputer, which consists of both a simulation component that generates electron microscopic data, and a learning component that trains to reconstruct the local electron density. } \label{fig:stemdl}
\end{figure}
With high performance communication fabric (e.g. InfiniBand) and software (e.g. NCCL enabled GPU Direct RDMA), data parallelism can achieve excellent scaling efficiency on HPC platforms. 
Figure ~\ref{fig:stemdl} shows an Exascale data-parallel deep learning application \cite{laanait2019exascale} on the Summit supercomputer, which tackles a long-standing inverse problem in atomic imaging (a.k.a phase problem) by deep learning reconstruction of the electron densities from microscopic images. About $93\%$ scaling efficiency has been achieved up to $27,000$ GPUs with a peak performance of $2.1$ EFLOPS in half precision.

\paragraph{Take Home Message.} 
Data-parallel training is the most popular distributed training methods thanks to it being model agnostic (CNN, RNN, or GAN) and simple to implement (framework built-ins, see Table~\ref{tab:dp}, or third party library plugins, e.g. Horovod). 
On the other hand, for applications with very a large model, a model may not be able to fit into a single device.
For application with very large input data, model accuracy is likely to suffer at large batch size. 
For those use cases, parallelism beyond data parallel paradigm needs to be explored, which will be discussed in the next section.     

\section{Model Parallelism}\label{sec:MP}

\begin{table}[]
\centering
\tbl{Evolution of model size and parallel framework, from data parallel training on computer vision models (typically of millions of parameters) to hybrid parallel training on large language models (up to billions of parameters).}{
\begin{tabular}{c|c|c|c}
\hline
Model       & Year & \begin{tabular}[c]{@{}c@{}}Number of\\ Parameters\\ (Billion)\end{tabular} & \begin{tabular}[c]{@{}c@{}}Parallel \\ Framework\end{tabular}        \\ \hline
ResNet50 \cite{resnet50} & 2015   & 0.025                                                                      & \begin{tabular}[c]{@{}c@{}}PyTorch\\ data parallelism\end{tabular}      \\ \hline
BERT-Large \cite{bert-large} & 2018 & 0.34                                                                       & \begin{tabular}[c]{@{}c@{}}TensorFlow  \\ data parallelism\end{tabular} \\ \hline
GPT-2 \cite{gpt2}    & 2019  & 1.5                                                                        & N/A                                                                  \\ \hline
Megatron-LM \cite{megatron} & 2019 & 8.3                                                                        & \begin{tabular}[c]{@{}c@{}}PyTorch\\ hybrid parallelism\end{tabular}    \\ \hline
T-NLG    \cite{deepspeed}  & 2020 & 17                                                                         & \begin{tabular}[c]{@{}c@{}}PyTorch\\ hybrid parallelism\end{tabular}    \\ \hline
\end{tabular}}
\label{tab:mp}
\end{table}

As illustrated in section~\ref{sec:WF}, another level of parallelism that can be explored is at the model level. 
This is useful when a model is too big to fit into the GPU memory such as state-of-the-art NLP models with billions of parameters \cite{megatron}, or when the input data is of very high dimension such as medical\cite{mesh-unet}, geospatial imaging or particle tracking~\cite{Farrell:2018cjr}.
As shown in Table~\ref{tab:mp}, the size of popular models has grown from millions to billions of parameters in just over 5 years.   

Given the execution dependency between layers of a neural network, simple device placement of different layers will not execute at the same time. There are two common implementations for model parallelism.
First, \textit{pipelining}, e.g. GPipe \cite{Huang2018GPipeET}, where a batch is divided into micro-batches and then different micro-batches at different layers can execute in parallel albeit at a reduced efficiency.
Second, \textit{tensor contraction}, e.g. Mesh-TensorFlow \cite{NIPS2018_8242}, where other dimensions of a typical input batch can be parallelized in addition to the batch dimension.  

In comparisons, the GPipe approach is more generic for sequential models but it suffers from low scaling efficiencies due to the so-called \textit{bubble} \cite{Huang2018GPipeET}, namely the idle time caused by the sequential dependency of the execution of a micro-batch among model layers, and the same large batch training issue as in data parallelism. On the other hand, the Mesh-TensorFlow method can be more efficient but it relies on customizing each operation in the model.        

In addition to augmenting parallel capabilities on top of existing popular frameworks such as TensorFlow and PyTorch, several notable efforts tackle the same issue from the traditional HPC perspective, including the Livermore big artificial neural network toolkit (LBANN \cite{lbann}), which provides model parallel acceleration via domain decomposition, and a Legion \cite{legion} based framework that uses task graph parallel strategy \cite{taskparallel}. 
Both domain decomposition and task graph are generic parallel computing techniques that go beyond distributed training and the scope of this chapter. 
These developments have seen adoptions typically in the HPC community. 

\begin{figure}[h!]
  \centering
  \includegraphics[scale=0.4]{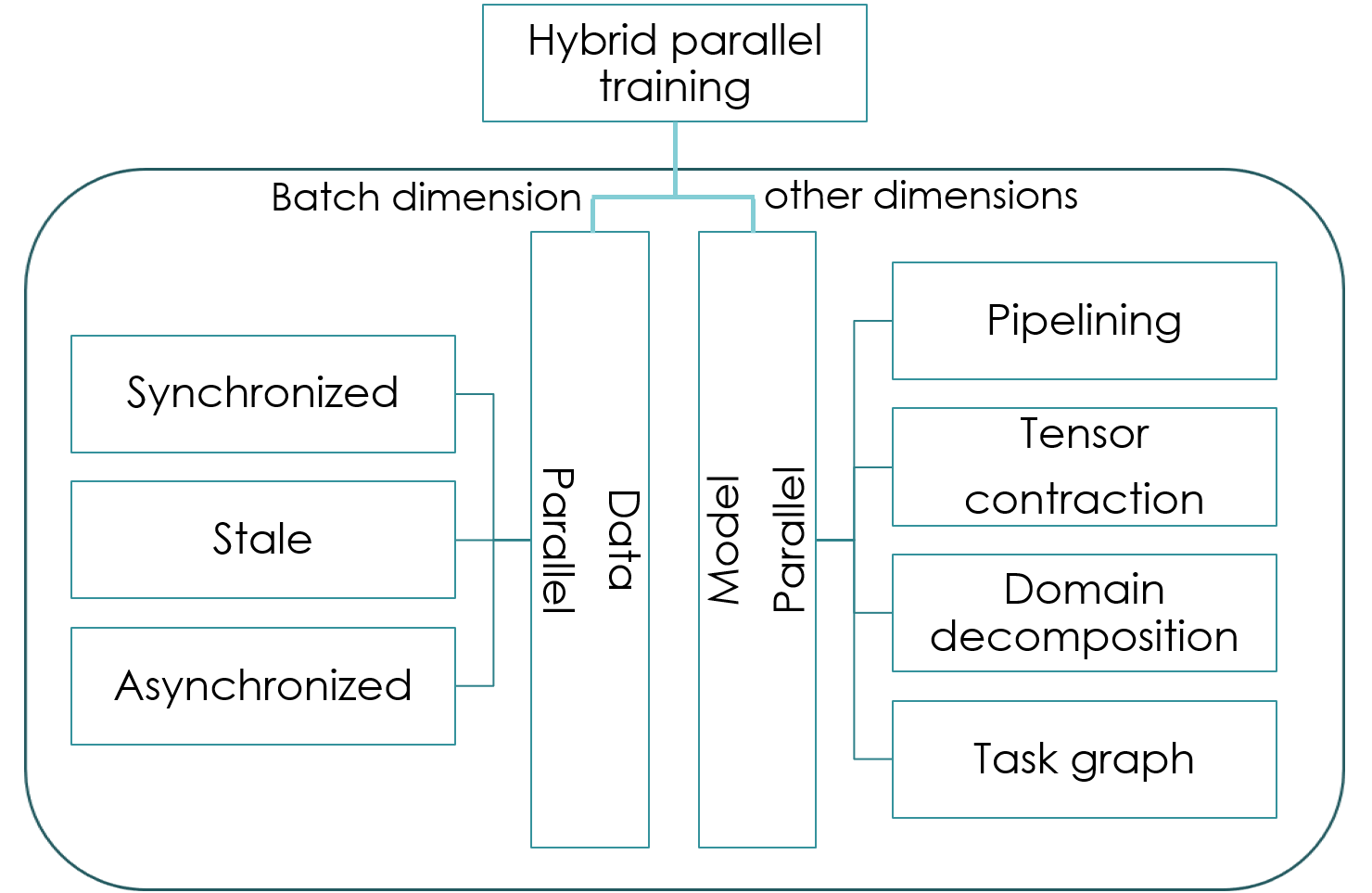}
  \caption{Hybrid parallel training scheme: data parallelism applied on the batch dimension and model parallelism on others. For data parallelism, depending on how often models are synchronized, it can be every step (synchronized), every few steps (stale), or asynchronized. For model parallelism, there are different implementations including pipelining~\cite{Huang2018GPipeET}, tensor contraction~\cite{mesh-unet}, domain decomposition~\cite{lbann}, and task graph~\cite{taskparallel}.} \label{fig:hybrid}
\end{figure}

In practice, the model parallelism alone usually will not scale due to the cross-node communication latency. Therefore, a hybrid scheme (see Fig.~\ref{fig:hybrid} and Table~\ref{tab:mp}) with data parallelism on the batch dimension and model parallelism on other dimensions strikes a better balance between scaling efficiency and model capabilities (larger model, faster convergence, etc) than that of data or model parallelism alone.   

\paragraph{Take Home Message.} 
Model parallelism is becoming more important as the model size and the input dimension grow. It further improves the distributed training efficiency at the cost of software complexity. 
So far, there is no common framework for model parallelism in the community and most developments are model specific. 
Model parallelism can be combined with the other means of parallelism presented so far and  is also fully orthogonal to hyper-parameter optimization discussed in the next section.

\section{Hyper-parameter optimization}\label{sec:HO}

As introduced in section~\ref{sec:WF}, the training procedure of a model, to obtain an optimal set of parameters $\theta^\star$ may depend on a set of hyper-parameters $h$.
The optimal value $h^\ast$ that yields the best of the optimal model parameters can be found by optimization of a figure of merit $\mathcal{F}$ --- usually the loss $\mathcal{L}$ (see section~\ref{sec:WF}).

When comparing the performance of two models obtained from two different sets $h_1$ and $h_2$, it is important to control whether the difference observed is relevant and significant.
If the training is done once on a training set from a unique set of initial parameters, and the model performance is evaluated on a fixed testing set, then there is an uncertainty related to these choices. 
The cross validation procedure provides a way to address this uncertainty and in particular to obtain a better estimator of the model performance.
The training set is split in $K$ parts --- or \textit{folds} --- $\left\{F_k\right\}$.
$K$ train/test pair can therefore be formed
\begin{equation}
    \left( \left\{ X_i \right\}_{i \notin F_k } , \left\{ X_i \right\}_{i \in F_k} \right)_k; k = 1...K.
\end{equation}
Running $K$ trainings one obtains a set of $K$ estimations of the figure or merit : $\{\mathcal{F}_k\}$.
The performance estimator $\overline{\mathcal{F}} = \mathbf{E}_k [\mathcal{F}_k]$ provides a better way to compare models trained under different settings.
The variance of the performance is an indicator of the instability of the model training when changing random seeds and training partition.
This might be considered also as a selection criterion for the model.
It should be noted that training over the $K$ folds can be done fully in parallel, and therefore offer an almost perfect scaling. 
Residual inefficiency can arise if the training over one fold requires significantly larger amount of resources than the others.

In most search strategies of the optimal hyper-parameters, the trainings with different parameter sets $h_i$ are done independently of each other and therefore offer another level of parallelism.
However, large fluctuation of resource needs can be expected for different parameter sets and some scaling bottlenecks may arise.
Other strategies~\cite{jaderberg2017population8638154} involve cross-talk between hyper-parameter sets, introducing some level of interlocking of processes, with the benefit of an improved search mechanism.

Strategies have been proposed for the optimization of hyper-parameters that do not involve the architectural aspect of the model (number of layers, neurons, ...).
In~\cite{jaderberg2017population8638154}, the model parameters and hyper-parameters are evolved at the same time over a population of models, effectively learning a scheduling of changes of hyper-parameters that favors the convergence of the model.
Conversely, some strategies concentrate only on the hyper-parameters driving the architecture of the model.
In~\cite{pasini2019greedy}, the architecture search starts with a simple model, and new layers are added in steps, on top of the best performing previous layer setup. 
This reduces the navigation of the large phase space of model architectures in which both the number of layers and number of neurons per layer are varied simultaneously, to a subset where only the size of the last layer is modified at once.

Strategies that aim at optimizing a generic set of hyper-parameters most commonly use a Bayesian optimization~\cite{snoek2012practical} or an evolutionary algorithm approach~\cite{7727192,Young:2017siw} as an efficient alternative to a simple grid search over the large space of hyper-parameters.
Bayesian optimisation with a Gaussian process (GP) assumption on the prior of the hyper-parameters is adequate if the prior is sufficiently regular to be modeled with a GP.
The optimization process is rather sequential --- the choice of a new set to evaluate is conditional to evaluation of all previous choices --- but some level of parallelism can be introduced to favor exploration.
The GP approximation can become computationally prohibitive if the number of steps is too large --- when the size of the hyper-parameters space is large for example.
The Bayesian approach is favorable when the user is limited in the number of usable parallel processes --- such as when making use of a small scale institutional cluster, or personal computer.
Evolutionary algorithms, on the other hand proceed with evaluating the performance (or fitness) of hyper-parameters sets in parallel --- as individuals of a population --- and synchronize all computing processes during population mutation and breeding.
This approach can offer a better exploration of hyper-parameters, at the cost of more parallel resources --- available as such on high performance computing (HPC) centers and large scale clusters.

\paragraph{Take Home Message.} 
Hyper-parameter search is an optimization problem on top of the model parameter optimization. 
A simple grid search algorithm becomes quickly prohibitive with the number of hyper-parameters. 
Bayesian optimisation may be recommended for an intermediate number of hyper-parameter, on intermediate scale computing cluster.
Evolutionary algorithms are adapted for large hyper-parameter space, and large scale computing clusters, such as HPC facility.

\section{Summary and Discussion}\label{sec:SD}

We have succinctly introduced the possible ways of running the computation for model training and optimization in parallel.
We reviewed the latest developments of each strategy, their strengths and weaknesses.
The otherwise highly nested loop involved in model training, K-folding, and hyper-parameter optimization, can be largely unfolded in parallel computations.
In the following, we provide remarks on key aspects of distributed training, and prospects in the field of particle physics.

\paragraph{Communication bottleneck.}
Similar to performance optimization of HPC applications, improving the scaling of distributed training is about finding the balance point between compute, I/O, and communication. 
Regardless of whether one uses data, model, or hybrid parallelism, most deep neural network applications can benefit from performance boost of the underlying communication protocol, since many such applications become network-bound at large scale (e.g. the message size is 100MB and 1.44GB for ResNet50 and BERT-large, respectively). 
It is reported \cite{deep-gradient-compression} that model accuracy can be maintained while the message size for gradients is being reduced by up to 2 order of magnitude via a combination of compression techniques (e.g. clipping, sparsification, accumulation, quantization).

\paragraph{Trading memory for computation.}
Since neural network training is usually performed on GPU devices and GPU memory is still a scarce resource, trading memory for computation is sometimes a necessary workaround.
It however requires significant engineering effort, similarly to implementing model parallelism.
One strategy is \textit{memory check-pointing} \cite{memory-checktpointing}, where only selected forward nodes in the computation graph are check-pointed while others are re-computed during the backward propagation for the gradient evaluation. 
This results into a $O(\sqrt{n})$ memory usage of a neural network model with $n$ nodes. 
Another strategy is via GPU memory management \cite{ibm-lms}, where inactive tensors are automatically transferred from GPUs to the host and vice versa. 
This is transparent to users and the added performance penalty is often tolerable.       
As reported in~\cite{oktay2020randomized}, a specific random sampling of the computation graph during the back-propagation phase allows for a reduced memory burden in computing the gradients.

\paragraph{Locality of resources.}
Depending on the resources, the optimal distributed training strategy may vary.
Computing nodes in HPC resources are often homogeneous and equipped with high performance interconnects, where the ring allreduce communication usually works the best but only half of the theoretical network bandwidth is achievable (see Sec.~\ref{sec:DD}).
Cloud resources, on the other hand, can be provisioned and are more suitable for a parameter-server distribution strategy.
BytePS \cite{byteps} is one such example to take advantage of cloud resources, and is reported  to perform better than the collective communication approach in a specific hardware setup, where total bandwidth of server nodes is no less than that of worker nodes. 
How to efficiently apply a similar parameter distribution approach to node-homogeneous HPC resources is not yet clear. 

In most of the discussion in this review, the input dataset is assumed to be located in local storage.
No consideration of data caching on the node or on the edge was done.
However, in particle physics, the available data may be distributed over multiple regions, and hence a specific data management protocol may be required.
The training computation could be done where the data is --- with implications on the software framework --- or the data would need to be moved and cached on the edge of the computing resource --- with implications on the data management system.

\paragraph{Running the models.}
The software used in particle physics to run model in production is actually not dictated by machine learning considerations, and restrictions might arise in how the models are encoded.
This requirement for compatibility of software might impose constraints on the framework used for distributed training, unless cross-framework conversion tools are kept extremely efficient.

Even in situations where the model is ran on remote platform~\cite{Duarte:2019fta,Krupa:2020bwg}, therefore loosening some of the software compatibility constraints, the model components might have to be post-processed in particular ways to fit on the remote hardware --- in particular on FPGAs.
As a consequence, some of these inference-oriented post-processing steps (pruning, quantization, ...) might need to be taken into account during the distributed training phase to ensure a maximally efficient algorithm in the end.

\paragraph{Framework consideration.}
In terms of software availability for distributed training, parameter and data parallelism are well supported both by frameworks (see Table~\ref{tab:dp}) and third-party plugins, e.g. Horovod \cite{horovod} for data parallelism, BytePS \cite{byteps} for parameter distribution, etc. 
For model parallelism, on the other hand, libraries are still in early stages, e.g. Mesh-TensorFlow \cite{NIPS2018_8242}, PyTorch-GPipe \cite{torchgpipe}, etc. 
There are also several hyper-parameter optimization libraries, like Ray Tune \cite{ray-tune}. 
What is lacking is a unified open software framework that can effectively explore all the parallelism in distributed training, and hence allow for an efficient use of computing resource and person power.

A particular aspect in particle physics is that budget constraints often lead research groups to develop their own software so that the support model stays within the community.
Although the problem of distributed training is not specific to high energy physics, some of the features could be specific, as explained above, and need sustainable support.
It is therefore of the utmost importance to have good feature support and maintenance in the \textit{framework-to-be-used} in particle physics.
%It is therefore necessary to have good integration of feature support and maintenance in the \textit{framework-to-be-used} in particle physics.

\paragraph{Final Remarks.}
As the high performance computing machines are entering the exascale era, distributed training and optimization of neural networks is one of the major areas that can harness this tremendous computing power. 
While there are many efforts exploring parallelisms from parameter distribution, data and model parallelism, to hyper-parameter optimization, the community is still in need of an overarching solution. 
One noteworthy work in this direction is a framework~\cite{taskparallel} based on the task graph, a parallel programming paradigm that is suitable for exascale. 
It can naturally combine the aforementioned distributed training strategies and hence enable an efficient end-to-end training workflow, although feature developments are lagging behind mainstream frameworks.

\section*{Acknowledgement}
We thank authors of other chapters, Vlad Oles and Feiyi Wang for feedback on our manuscript.
J.Y. is  supported  by  the  U.S.  Department  of  Energy,  Office  of  Science, National Center for Computational Sciences. This  research  used  resources  of  the  Oak  Ridge  Leadership  Computing  Facility,  which  is supported by the Office of Science of the U.S. Department of Energy under Contract No. DE-AC05-00OR22725.
J-R.V. is partially supported by the European Research Council (ERC) under the European Union's Horizon 2020 research and innovation program (grant agreement n$^o$ 772369) and  by the U.S. Department of Energy, Office of Science, Office of High Energy Physics under award numbers DE-SC0011925, DE-SC0019227 and DE-AC02-07CH11359.

\bibliographystyle{ws-rv-van}
\bibliography{dist-train}
\end{document}